\documentclass[10pt,twocolumn,letterpaper]{article}

\usepackage{iccv}
\usepackage{times}
\usepackage{epsfig}
\usepackage{graphicx}
\usepackage{amsmath}
\usepackage{amssymb}

\usepackage[caption=false]{subfig}
\usepackage{multirow}
\usepackage{animate}
\usepackage{textcomp}

\usepackage[breaklinks=true,bookmarks=false]{hyperref}

\iccvfinalcopy 

\def\iccvPaperID{14} 

\ificcvfinal\fi

\begin{document}

\title{Multitask Learning to Improve Egocentric Action Recognition}

\author{
Georgios Kapidis\textsuperscript{1,2},~ 
Ronald Poppe\textsuperscript{2},~
Elsbeth van Dam\textsuperscript{1},~
Lucas Noldus\textsuperscript{1},~
Remco Veltkamp\textsuperscript{2}\\
\textsuperscript{1} Noldus Information Technology, Wageningen, The Netherlands\\
\tt\small{\{georgios.kapidis, elsbeth.vandam, lucas.noldus\}@noldus.nl}\\
\textsuperscript{2} Department of Information and Computing Sciences, Utrecht University, Utrecht, The Netherlands\\
\tt\small{\{g.kapidis, r.w.poppe, r.c.veltkamp\}@uu.nl}
}

\maketitle
\ificcvfinal\thispagestyle{empty}\fi

\begin{abstract}
In this work we employ multitask learning to capitalize on the structure that exists in related supervised tasks to train complex neural networks. It allows training a network for multiple objectives in parallel, in order to improve performance on at least one of them by capitalizing on a shared representation that is developed to accommodate more information than it otherwise would for a single task. We employ this idea to tackle action recognition in egocentric videos by introducing additional supervised tasks. We consider learning the verbs and nouns from which action labels consist of and predict coordinates that capture the hand locations and the gaze-based visual saliency for all the frames of the input video segments. This forces the network to explicitly focus on cues from secondary tasks that it might otherwise have missed resulting in improved inference. Our experiments on EPIC-Kitchens and EGTEA Gaze+ show consistent improvements when training with multiple tasks over the single-task baseline. Furthermore, in EGTEA Gaze+ we outperform the state-of-the-art in action recognition by 3.84\%. Apart from actions, our method produces accurate hand and gaze estimations as side tasks, without requiring any additional input at test time other than the RGB video clips.
\end{abstract}

\section{Introduction}
\label{sec:intro}

Human activity recognition from video is a growing field of computer vision that promises real-time and large-scale behavior recognition and automated analysis. Activity recognition applies to both the third and first-person vision domains, incorporating the distinct visual characteristics of each case. Third-person videos tend to capture the full range of motions of the human body from a static point of view. The viewing angle in egocentric videos matches that of the human performing the activity, providing a unique, moving perspective of the scene \cite{kanade_first-person_2012}. At the same time egocentric videos usually offer a clear view of the camera wearer's hands \cite{fathi_understanding_2011}, which in many cases are essential for the execution of an activity. An outlook of the objects manipulated by human hands promises additional cues about the performed activity culminating to improved recognition performance \cite{fathi_understanding_2011, li_delving_2015, baradel_object_2018}.

\begin{figure}[!tp]
    \centering
    
    \begin{animateinline}[poster=0, autoplay, loop, controls]{2}
    \multiframe{16}{I=0+1}{
        \begin{tabular}{@{\hspace{0pt}}c@{\hspace{0.02\linewidth}}c@{\hspace{0pt}}}
            \includegraphics[width=0.48\linewidth]{latex/figures/1/nh_\I} & 
            \includegraphics[width=0.48\linewidth]{latex/figures/1/h_\I}
        \end{tabular}
    } 
    \end{animateinline}
    \caption{Visualizing the class activation maps \cite{stergiou_saliency_2019} for an instance of class `open' from EPIC-Kitchens \cite{damen_scaling_2018}. Left: Multi-Fiber Network (MFNet) \cite{chen_multi-fiber_2018} trained end-to-end for the single task of classifying short clips into actions. Right: MFNet trained to additionally predict one (x,y) coordinate for each hand. 
    Training with the hand coordinates as the extra task leads to a greater inclusion of the right hand area into the class activation map.}
    \label{fig:1}
\end{figure}

Recent methods for video activity recognition employ convolutional (CNN) and recurrent (RNN) deep neural network structures to capture the information from RGB images or video frames, regardless of the viewing perspective \cite{karpathy_large-scale_2014, donahue_long-term_2017}. More recent approaches use 3D convolutions \cite{carreira_quo_2017} to incorporate the temporal information that resides in frame sequences. Occasionally, they are enhanced with attention schemes \cite{girdhar_attentional_2017} to select specific features or frames as more informative for an activity. In order to expand the feature space, the network input can be augmented with additional data modalities. These can be optical flow \cite{simonyan_two-stream_2014}, depth \cite{tang_action_2017} or input segmentation masks around interesting areas \cite{tu_multi-stream_2018}. The information from them aims to guide the network towards learning more activity-specific features that it might otherwise have missed. In order to incorporate the supplementary inputs, networks comprise multiple streams, i.e. parts of the original structure are copied in the number of modalities, trained individually (e.g. \cite{wang_temporal_2016}) and their results are combined at a later stage. The multi-stream approach is associated with the combination of the individual feature sets towards an extended and more expressive representation from which activities are inferred. 

A related but fundamentally different concept that we employ for this work is that of Multitask Learning (MTL) \cite{caruana_multitask_1997}. The idea behind MTL is to train a neural network with multiple related objectives (tasks) while sharing as much as possible of a common network structure \cite{caruana_multitask_1997}. Branch diversification occurs only for the task-specific output layers and there are as many output layers (branches) as there are learnable tasks spawning from the main network block, increasing the dimensionality of the output. 

MTL is conceptually the opposite from multi-stream approaches, since the additional information is not used as input but is expressed as the output of the network and is only required for supervision. The significant merit of MTL over multi-stream methods is that the additional information is only needed during training and what would otherwise come from the additional input modalities is already incorporated in the network weights at test time. For example, in the video domain, the input of a network remains the same set of RGB images regardless of the number of tasks.

The premise of MTL is that by combining the objectives of related tasks in the same network, we can benefit from their structural commonalities. This is the case because the weights of the shared network block aim to jointly encapsulate each task's representation requirements. When these are complementary, they enhance the inputs of the task-specific output layers. Then, inference can be improved for all or some of them or just the one that we focus on the most, by saving its best performing weights \cite{caruana_multitask_1997}. 

In this work, we utilize MTL to improve action classification performance in egocentric videos. We are motivated by the idea that hands are critical for the comprehension of egocentric actions but it remains difficult for networks to capture this delicate motion information. In Figure~\ref{fig:1}, we show that by having the network learn hand regions explicitly as an extra task in addition to the actions, we steer it to produce activation maps that cover the corresponding hand areas to a greater extent. Eventually, incorporating these areas also improves the action classification results. 

We experiment with egocentric datasets EPIC-Kitchens \cite{damen_scaling_2018} and EGTEA Gaze+ \cite{li_eye_2018} by explicitly utilizing the location of hands, gaze and other signals towards actions. We leverage the motion and visual attention information that is present in the hand movements and the gaze of the camera wearer, respectively, which have proven descriptive for predicting egocentric actions on their own \cite{fathi_understanding_2011, li_delving_2015, kapidis_egocentric_2019}. In addition, we show that when complementary classification tasks are added during training, performance improves further. 

The contributions of this paper are the following:
\begin{itemize}
    \item An MTL scheme that extends 3D CNNs  \cite{chen_multi-fiber_2018} and functions with an arbitrary number of output tasks.
    \item Experiments demonstrating that MTL improves on egocentric action classification over singe-task learning (STL) baselines without requiring any additional information at test time, other than the input video.
    \item Experiments generalizing our MTL scheme to a number of related classification and coordinate estimation tasks that improve egocentric action classification.
\end{itemize}

In Section~\ref{sec:related_work} we review recent work on action classification and MTL. In Section~\ref{sec:methodology} we develop our MTL pipeline for an arbitrary number of tasks. In Section~\ref{sec:experiments} we document our experiments on two egocentric video datasets. In Section~\ref{sec:discussion} we discuss our findings and in Section~\ref{sec:conclusions} we conclude.

\section{Related work}
\label{sec:related_work}
In Section~\ref{sec:rel_feats} we discuss feature-based egocentric action recognition approaches. We continue with the more recent deep network advances for activity recognition from the third person perspective in Section~\ref{sec:rel_3rd-adv} and their expansion into egocentric in Section~\ref{sec:rel_1st-adv}. In Section~\ref{sec:rel_mtl} we discuss related work from the perspective of MTL.

\subsection{Feature-based egocentric action recognition} 
\label{sec:rel_feats}
The hands, the manipulated objects, ego-motion and their interrelationships have been established as some of the most prominent characteristics for egocentric action recognition \cite{fathi_understanding_2011, spriggs_temporal_2009, kanade_first-person_2012}. In this observation lies the origin of the hand-crafted feature approaches that prevail in earlier works in egocentric vision.

Fathi~\etal~\cite{fathi_understanding_2011} use hand and object segmentations to infer actions and based on feedback from the latter improve the initial hand and object detections. The importance of the detected objects and the interactions between them for activities is highlighted in \cite{pirsiavash_detecting_2012}, where activity recognition improves using additional information about objects being either passive or actively engaged with in the scene. In \cite{fathi_learning_2012}, the gaze of the camera wearer is used to define the salient areas in first-person views, recognizing that egocentric actions are further correlated with modalities that describe human attention in the video. Global and local motion are considered in \cite{ryoo_first-person_2013} to produce features that describe egocentric actions. Feature-based egocentric action recognition is analyzed in \cite{li_delving_2015} where the importance of motion and object cues, hand, head movements and gaze are evaluated in various combinations. A review of these approaches appears in \cite{nguyen_recognition_2016}. In our work, we do not use an explicit feature-based representation of modalities for the input, but use data from these modalities as supervision to learn actions.

\subsection{Advances in third-person activity recognition}
\label{sec:rel_3rd-adv}
Recent work in third-person vision \cite{poppe_survey_2010, stergiou_understanding_2018} has seen the successful employment of deep network approaches. We highlight the work of Karpathy~\etal~\cite{karpathy_large-scale_2014} who use 2D CNN architectures to classify video frames and in order to incorporate information from multiple frames, explore various fusion schemes to enhance the classification output. Other approaches include the use of two-stream networks \cite{simonyan_two-stream_2014, wang_temporal_2016, gkioxari_finding_2015} that capture appearance and motion in images with spatial and motion streams trained on single or multiple frames concurrently. Further attempts to take advantage of the temporal consistency in videos consider recurrent units following frame-wise feature-extracting CNNs \cite{donahue_long-term_2017, li_videolstm_2018}. More recent approaches in video activity recognition use 3D CNNs \cite{tran_learning_2015, carreira_quo_2017, tran_closer_2018, chen_multi-fiber_2018}. Here, video frames are modelled as a result of convolutional kernels being learned not only on the spatial dimension of images, but also on the changing pixel values in frame sequences. We find that using a 3D CNN to capture patterns in the temporal dimension also works well for egocentric action recognition.

\subsection{Advances in first-person activity recognition}
\label{sec:rel_1st-adv}
\paragraph{Spatial networks} A large volume of work in egocentric vision already incorporates these advances. A CNN appearance feature extractor from images is used to categorize egocentric actions in \cite{ryoo_pooled_2015}. In \cite{wray_sembed_2016}, CNN features from RGB images are used to produce embeddings that are semantically linked among videos and are used as the basis to model relationships between objects and actions and classify them. In \cite{ma_going_2016}, a two-stream network is trained to capture appearance and motion features. The appearance stream is pretrained for hand segmentation and is finetuned for object localization, to find the interesting regions of the image. The motion stream is trained on optical flow and both feature representations are combined with late fusion to predict short-term actions, objects and activities. Furthermore, in \cite{singh_first_2016}, a three-stream architecture trained on egocentric features from hand masks, head motion and saliency maps for the first stream, and raw RGB images and optical flow maps for the other two is adapted to recognize actions. We also highlight the improvements from specific egocentric cues such as hand movements and gaze-based visual saliency on top of the image-based features, but from a contrasting perspective. Instead of using them as input, we consider them as additional learnable tasks with the advantage of not requiring the extra information at test time.
\paragraph{Spatio-temporal with recurrence} Approaches utilizing recurrent networks following 2D CNNs include \cite{vaca-castano_improved_2017, huang_egocentric_2018, song_multimodal_2016, kwon_first_2018}. CNN features are propagated to recurrent units for temporal action proposal generation \cite{huang_egocentric_2018}, action related scene identification \cite{vaca-castano_improved_2017} and action recognition \cite{song_multimodal_2016, kwon_first_2018}.

\paragraph{Recurrent with attention} The temporal aspect of videos is further studied with recurrent attention mechanisms \cite{bolanos_egocentric_2018, shen_egocentric_2018, sudhakaran_attention_2018, sudhakaran_lsta_2019, li_eye_2018, furnari_what_2019, wu_long-term_2019} that act to find the most informative parts in images (spatial attention) or the most informative frames throughout videos (temporal attention). An encoder-decoder scheme is described in \cite{bolanos_egocentric_2018} for textual description of videos. Here, the current event's frames are encoded into CNN features and modelled temporally with LSTM. From the current and previous step's embedding, an attention mechanism selects the features that will be decoded as the optimal textual description of the current activity. The attention mechanism in \cite{shen_egocentric_2018} focuses on the frames that carry the action specific information by learning the associations between the input gaze, the detected objects and the segmented hands. The combined focus on these regions allows the network to discard redundant frames of the input video segment that would otherwise obfuscate the prediction task. Spatial attention is considered in \cite{sudhakaran_attention_2018} where the important regions from every frame are given as input to an LSTM for action recognition, whereas in \cite{sudhakaran_lsta_2019} spatial attention is further correlated sequentially through continuous frames. In \cite{wu_long-term_2019}, attention is based on video specific spatial features (such as person detections in third-person videos and objects with motion for egocentric), which are calculated intermittently over the course of a video. These temporally examined spatial features introduce past information to predict the current action. In \cite{furnari_what_2019}, the attention mechanism weighs the importance of the input modalities to select optimal features for action anticipation and recognition. In contrast to explicit attention mechanisms, we are using the additional supervisory tasks to enhance the representation in deeper layers by incorporating information from all and learning them together, thus inciting the network to acquire universally useful features. 

\subsection{Multitask learning}
\label{sec:rel_mtl}
Training a network for multiple tasks jointly has been shown to improve the performance on all of them or at least the main task, as long as they share a conceptual similarity \cite{caruana_multitask_1997} or they are not competing \cite{sener_multi-task_2018}. In \cite{sener_multi-task_2018}, the problem of loss function weighing is analyzed, to train for multiple tasks efficiently with an optimization scheme that searches for the optimal set of network parameters for the best trade-off among tasks. Instead, we weigh all tasks equally with our focus on properly selecting related tasks in advance. 

MTL has been applied to computer vision problems such as joint object and action detection \cite{kalogeiton_joint_2017}, object detection and segmentation \cite{he_mask_2017} and boundary, surface normal and saliency estimation together with object segmentation and detection \cite{kokkinos_ubernet_2017}. In \cite{zamir_taskonomy_2018}, the relationship between tasks is modelled in a latent space to transfer knowledge between them and reduce the number of required training samples. MTL in egocentric vision appears in \cite{baradel_object_2018, ma_going_2016, li_eye_2018, huang_mutual_2019, nakamura_jointly_2017, sudhakaran_lsta_2019}. In \cite{baradel_object_2018}, from an RGB video input, multiple network branches learn activities, object proposals and segmentations, but with large parts of the network trained independently. In \cite{ma_going_2016}, an object and an action learning task are combined to produce an activity prediction (as a combination of the two), but the network layers except the last one are not sharing parameters. In \cite{li_eye_2018}, the network learns a gaze map which is used to pool from the activations of the final feature map for actions, thus the two tasks affect each other. Similarly, in \cite{huang_mutual_2019}, a prior gaze estimation is used to influence the action prediction, which in turn affects the final gaze prediction. Training takes place jointly, but internally, each network part is deployed for a specific task, without parameter sharing. Another example of multitask learning in egocentric vision is from \cite{nakamura_jointly_2017} with joint learning of activities and energy expenditures from video. Still, the input to the network is multimodal (video and accelerometer signals) and each stream is trained individually with parameter sharing only during a late fusion stage. Finally, in \cite{sudhakaran_lsta_2019} the action task is augmented by a verb and a noun learning task similar to ours, however, the bias of the action classifier is applied to the secondary tasks and alters their output explicitly. In this work, we do not attempt to influence task outputs explicitly, but employ parameter sharing in every network layer, down to the last step before the final output, as a way to induce implicit information sharing between them.

\section{Methodology}
\label{sec:methodology}

In Section~\ref{sec:meth:net_adapt}, we describe the process to adapt a network from single to multitask and in Section~\ref{sec:meth:task_out} the output layers with their individual loss functions for the tasks we consider. In Section~\ref{sec:meth:coord_pred}, we discuss the details of the coordinate prediction layer and its application to 3D CNNs to model the progression of movements through time.

\subsection{Multitask network structure}
\label{sec:meth:net_adapt}

\begin{figure}[!ht]
    \centering
    \includegraphics[width=\columnwidth]{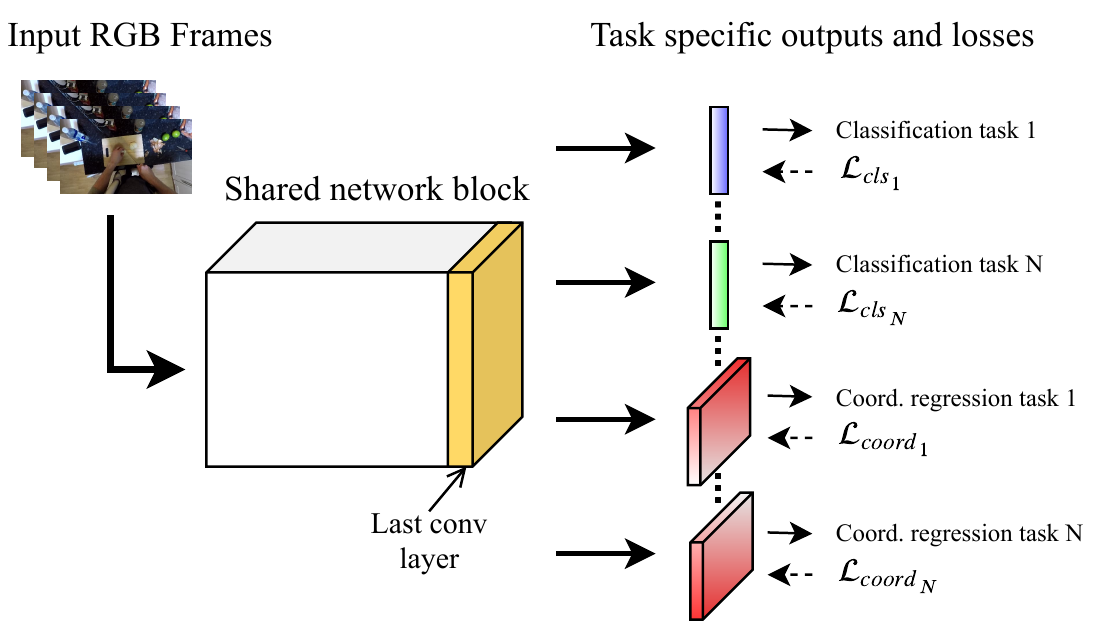}
    \caption{MTL Network Structure. The shared network block can be any convolutional network that extracts features from the input. Each task-specific output layer is plugged to the `last conv layer'. They are independent from the others and their parameters are trained individually. However, all output layers use the feature representation produced by the shared part of the network as input.}
    \label{fig:2}
\end{figure}

In Figure~\ref{fig:2} we visualize our network architecture for multitask training. The backbone of the network is a feature extractor (the shared network block), upon which the task-specific output layers are attached. We represent the feature extracting network block with  \(g(x;\theta)\), where \(x\) is an input data point from input space \(\mathcal{X}\) and \(\theta\) are the parameters of \(g\). For each task \(t\), we define an output function \(f_{t}(g(x);\theta_{t})\), where \(\theta_{t}\) are the parameters of the task-specific layer and \(t\in \mathcal{T}\), with \(\mathcal{T}\) being the set of tasks. For this work, function \(g\) is approximated with a 3D Convolutional Neural Network and the input space \(\mathcal{X}\) is defined as a set of RGB images sampled from a video clip.

\subsection{Task-specific output layers}
\label{sec:meth:task_out}

In order to train the network, we formulate the loss function based on the number and types of tasks it encapsulates. We perform MTL with two types of tasks: classification and coordinate regression. For a classification task \(i\), \(\mathcal{L}_{cls_{i}}\) is the categorical cross-entropy loss. For a coordinate regression task \(j\), \(\mathcal{L}_{coord_{j}}\) is defined as the Differentiable Spatial to Numerical Transform (DSNT) loss from \cite{nibali_numerical_2018}, explained in Section \ref{sec:meth:coord_pred}. The full loss function \(\mathcal{L}\) is defined as
\begin{equation}
    \label{eq:1}
    \mathcal{L} = \sum_{i} \mathcal{L}_{cls_{i}} + \sum_{j} \mathcal{L}_{coord_{j}}.
\end{equation}

During training the value of \(\mathcal{L}\) is not propagated through each task-specific layer, but each task layer \(t\) produces gradients with respect to its individual loss, hence its parameters \(\theta_{t}\) are not affected by the remaining tasks. Finally, the gradients from all the output layers are summed and backpropagated through \(g\).

\subsection{Coordinate prediction}
\label{sec:meth:coord_pred}

Our approach for predicting coordinates stems from the numerical coordinate regression layer introduced in \cite{nibali_numerical_2018}. It enables a 2D CNN to output an \((x, y)\) coordinate without using a fully connected layer, thus ensuring spatial invariance in the predicted coordinate \cite{nibali_numerical_2018}. Instead, it relies on an additional convolutional layer that predicts a heatmap \(Z\) of shape \(m\times n\). The softmax activation is applied on \(Z\), such as \(\hat{Z}=\sigma(Z)\), to create a 2D probability distribution,
which is passed through the Differentiable Spatial to Numerical Transform (DSNT) layer to become a coordinate. 

In DSNT, \(\hat{Z}\) is discretized by calculating its Frobenius inner product for each dimension, against two uniformly distributed vectors with values in [-1, 1], shaped \(m\times1\) and \(1\times n\) respectively and copied over their singular dimension ($n$ and $m$ times) to become matrices the shape of \(\hat{Z}\). The output value of the Frobenious inner product for each matrix is the respective coordinate value with sub-pixel precision in the range [-1, 1]. This process preserves differentiability through the layer and allows gradient flow from a loss function directly associated with the error in coordinate space instead of the error in heatmap space.  

The coordinate loss \(\mathcal{L}_{coord}\) is the Euclidean distance between the predicted (\(c_{p}\)) and the expected (\(c_{gt}\)) coordinate with an added regularization factor \(\lambda=0.5\) to smooth the gradients around the prediction, i.e.
\begin{equation}
    \label{eq:2}
    \mathcal{L}_{coord}=\lambda\mathcal{L}_{euc}(c_{p}, c_{gt}) + (1-\lambda)\mathcal{L}_{reg}(\hat{Z}),
\end{equation}
where the Euclidean loss is
\begin{equation}
    \mathcal{L}_{euc}=\lVert(c_{p},c_{gt})\rVert_{2}
\end{equation} 
with \(c_{p}=DSNT(\hat{Z})\) and the regularization loss is
\begin{equation}
    \mathcal{L}_{reg}(\hat{Z})=\mathcal{L}_{JS}(\hat{Z},c_{gt})=JS(\hat{Z}\parallel\mathcal{N}(c_{gt},\sigma^{2}))
\end{equation}
based on the Jensen-Shannon divergence.

In order to successfully apply this coordinate regression layer in our setup, we need to account for the output dimensions of the last convolutional layer of the 3D CNN. In the 2D case that is \(B\times C\times m\times n\) with \(B\) the batch size and \(C\) the channel dimension. In the 3D case the output shape extends to \(B\times C\times l\times m\times n\) with \(l\) the added temporal dimension due to the 3-dimensional input. This leads to having \(l\) heatmaps \(Z\) as well as \(l\) coordinate losses (instead of \(1\)) for an RGB clip, which are averaged over the temporal dimension to avoid huge gradients.

\section{Experiments}
\label{sec:experiments}
In this section we describe our experiments for various task combinations. Besides having improved performance on action recognition we produce accurate detectors for hands and gaze without complicated modules other than the DSNT layers for the coordinate prediction tasks. Finally, we compare against the state-of-the-art.

\subsection{Datasets}
\label{sec:exp:datasets}

We use EPIC-Kitchens \cite{damen_scaling_2018} and EGTEA Gaze+ \cite{li_eye_2018} for our experiments. EPIC is a collection of 432 videos by 32 participants, performing kitchen-related activities in their homes. We use the publicly available training split which consists of 272 videos by 28 participants (28,470 action segments). It is annotated with start and end times and 125 verb and 352 noun class labels, the valid combinations of which are 2,521 actions. We partition the fully annotated training split into custom training and validation splits with participants 1-29 used for training (26,375 clips) and 30-31 for validation (2,095 clips). EGTEA consists of 86 videos of 32 people in 7 scenarios of food preparation activities in kitchens. The videos are cropped into 10,321 clips based on action segment annotations. The dataset comes with three predefined training and testing splits, with the first one comprising 8,299 and 2,022 clips, respectively. We use this split unless stated otherwise, to train and evaluate at the clip level \cite{li_eye_2018}. Each clip is labeled from 19 verbs and 53 nouns and their 106 valid action combinations in the dataset. In addition, the dataset is complemented with a gaze annotation for every video frame, which consists of an \((x,y)\) coordinate and its type (fixation, saccade or unknown).

\paragraph{Hand locations}
One of the tasks we consider for MTL is hand coordinate prediction. Similar to gaze annotations, this task requires a supervision signal for the hand locations on each frame. To accommodate our experiments we use the egocentric hand detection, tracking and identification algorithm from \cite{kapidis_egocentric_2019} to produce hand location information for every video frame. It uses \cite{redmon_yolov3_2018} to detect hand bounding boxes, \cite{bewley_simple_2016} to track them through time and hand-crafted priors to remove false-positives and identify between left and right hands. We further modify \cite{kapidis_egocentric_2019} to track the top right area of the left hand bounding box and the top left area of the right hand leading to coordinates that more accurately pinpoint the hands instead of the forearms.  

\subsection{Training and evaluation}
\label{sec:exp:train_eval}

For the shared network block of Figure~\ref{fig:2} we employ the Multi-Fiber Network (MFNet) \cite{chen_multi-fiber_2018}. It contains 3D convolutional layers in its structure to capture spatio-temporal information from frame sequences and uses a relatively low number of parameters (8M) and computational resources (11.1 GFLOPs), which leads to an efficient training scheme for large video datasets. For all our experiments we use weights pretrained on Kinetics \cite{carreira_quo_2017} and retrain the full network structure end-to-end on the respective dataset. 

We train with a triangular cyclical learning rate \cite{smith_cyclical_2017} policy that shifts learning rate from \(5\times10^{-4}\) to \(5\times10^{-3}\) and back in 20 epochs. For optimization we use stochastic gradient descent with Nesterov momentum \((0.9)\) and weight decay (\(5\times10^{-4}\)). We input a sequence of 16 frames, randomly scaled to \(256\times256\) and cropped to \(224\times224\). Frames are uniformly sampled from a 32-frame window that starts at a random point of an action video segment and does not exceed its last frame. Batch size is 32 for our setup with two Nvidia 1080Ti GPUs, training lasts for 60 epochs and results are reported for the best performing epoch for the main task (early stopping). 
To evaluate we sample uniformly 16 frames from a window of 32, centered around the temporal center of an action segment. We resize them to \(256\times256\) and use the \(224\times224\) center crop as the network input.

\begin{table*}[!ht]
    \centering
    \begin{tabular}{l|@{\hspace{2pt}}c@{\hspace{5pt}}c@{\hspace{5pt}}c@{\hspace{2pt}}|@{\hspace{2pt}}c@{\hspace{5pt}}c@{\hspace{5pt}}c@{\hspace{2pt}}|@{\hspace{2pt}}c@{\hspace{5pt}}c@{\hspace{5pt}}c@{\hspace{2pt}}|@{\hspace{2pt}}c@{\hspace{5pt}}c@{\hspace{5pt}}c@{\hspace{2pt}}}
        \multicolumn{1}{l}{} & \multicolumn{3}{c}{Top1 Acc. (\%)} & \multicolumn{3}{c}{Top5 Acc. (\%)} & \multicolumn{3}{c}{Avg class Prec. (\%)} & \multicolumn{3}{c}{Avg class Rec. (\%)} \\
        Tasks & Actions & Verbs & Nouns & Actions & Verbs & Nouns & Actions & Verbs & Nouns & Actions & Verbs & Nouns \\
        \hline
        V & - & 48.57 & - & - & 78.32 & - & - & 34.39 & - & - & 25.29 & - \\
        V + H & - & \textbf{49.31} & - & - & \textbf{78.80} & - & - & 29.85 & - & - & 25.68 & - \\
        V + N + H & - & 47.47 & \textbf{27.6} & - & 78.37 & \textbf{51.19} & - & 27.80 & 21.43 & - & 23.61 & 18.80 \\ 
        \hline
        A & 18.48 & - & - & 36.20 & - & - & 2.76 & - & - & 2.67 & - & - \\
        A + V + H & 18.58 & 49.05 & - & \textbf{38.82} & 78.75 & - & 2.89 & 28.43 & - & 2.87 & 23.23 & - \\ 
        A + V + N + H & \textbf{19.29} & 48.9 & 27.27 & 35.91 & 78.18 & 47.85 & 3.25 & 29.31 & 22.68 & 3.04 & 24.03 & 17.84\\ 
    \end{tabular}
    \caption{Multitask learning results on EPIC-Kitchens. The first column shows the trained tasks for a model: Actions (A), Verbs (V), Nouns (N) and Hands (H). We report Top1 and Top5 accuracy on our validation set. Average class precision and recall are reported for many-hot verbs, nouns and actions. Many-hot verbs and nouns have more than 100 instances in our training set. Many-hot actions are the valid combinations of many-hot verbs and nouns with at least one instance in the training set, following \cite{damen_scaling_2018}.}
    \label{tab:1:abl_epic}
\end{table*}

\subsection{Results on EPIC-Kitchens}
\label{sec:exp:epic}

Our results on EPIC-Kitchens are summarized in Table~\ref{tab:1:abl_epic}. Initially, we train the STL baseline with verbs for supervision. Then, we combine verbs with nouns and hands as separate tasks. Training for hands together with verbs (V+H) increases Top1 accuracy to 49.31\% (+0.75). Adding nouns (V+N+H) harms verb Top1 by 1.1\% but produces our best performing noun classifier.

In the EPIC-Kitchens literature \cite{damen_scaling_2018} verb and noun predictions are combined following their individual inference stages and are later synthesized into an action prediction. In our MTL scheme we train for the action task explicitly, i.e. the 2,521 valid verb and noun combinations. Having actions, verbs and hands for supervision (A+V+H) leads to 49.05\% for verbs, improving on the STL baseline by 0.48\% and additionally using the nouns (A+V+N+H) still improves from STL (+0.33). However, both cases do not improve as much as with only the hands, implying a conflict between the extra tasks. On the other hand, if we consider actions as the main task, the addition of verb, noun and hand learning tasks will only improve on the action STL baseline reaching 19.29\% (+0.81 from A and +0.71 from A+V+H).

\subsection{Results on EGTEA Gaze+}
\label{sec:exp:gtea}

\begin{table*}[!ht]
    \centering
    \begin{tabular}{l|ccc|ccc|ccc}
         \multicolumn{1}{l}{} & \multicolumn{3}{c}{Top1 Acc. (\%)} & \multicolumn{3}{c}{Top5 Acc. (\%)} & \multicolumn{3}{c}{Mean Cls Acc. (\%)} \\
        Tasks & Actions & Verbs & Nouns & Actions & Verbs & Nouns & Actions & Verbs & Nouns \\
        \hline
        A & 63.75 & - & - & 91.05 & - & - & 55.35 & - & - \\
        A + V & 67.80 & 79.03 & - & 91.89 & 99.41 & - & 59.15 & \textbf{79.44} & - \\
        A + V + N & 68.00 & 78.98 & 78.93 & 91.94 & 99.31 & 96.24 & 59.67 & 78.24 & 72.06 \\
        A + G & 66.59 & - & - & 91.54 & - & - & 59.44 & - & - \\
        A + H & 67.46 & - & - & 91.99 & - & - & 59.78 & - & - \\
        A + G + H & 66.12 & - & - & 90.54 & - & - & 58.91 & - & - \\
        A + V + N + G & 68.74 & 78.14 & \textbf{79.13} & 91.59 & 99.41 & \textbf{96.54} & 60.34 & 79.29 & 72.03 \\
        A + V + N + H & 68.20 & \textbf{79.18} & 77.94 & \textbf{92.24} & \textbf{99.51} & 96.34 & 60.13 & 79.34 & 71.1 \\
        A + V + N + H + G & \textbf{68.99} & 79.08 & 79.03 & 91.74 & 99.26 & 96.39 & \textbf{61.40} & 77.40 & \textbf{72.49}
        
    \end{tabular}
    \caption{Multitask learning results on EGTEA Gaze+. The first column shows the names of the supervised tasks: Actions (A), Verbs (V), Nouns (N), Gaze (G) and Hands (H). We report Top1, Top5 and Mean class accuracy on the first split of the EGTEA Gaze+ test set.}
    \label{tab:2:abl_gtea}
\end{table*}

In Table~\ref{tab:2:abl_gtea} we delineate our results as the network moves from one to multiple tasks in the EGTEA Gaze+ dataset. We establish the action STL baseline (A) at 63.75\% Top1 accuracy. Next, we train using additional supervision from verbs and nouns (A+V) and (A+V+N) and reach 67.80\% (+4.05) and 68.00\% (+4.25), respectively. For further experiments we utilize coordinate regression layers to train on gaze points and hand tracks. We see that with either task we improve in both Top1 and mean class accuracy over STL; A+G is 66.59\% (+2.84) and A+H is 67.46\% (+3.71). Further improvements stem from training for all classification tasks together with gaze or hand prediction. A+V+N+G reaches Top1 68.74\% (+4.99) and A+V+N+H is 68.20\% (+4.45). The attempt to combine gaze and hand coordinate regression tasks only with actions shows that the two coordinate tasks are competing to influence the shared representation and have the smallest improvement over the STL baseline with A+G+H Top1 at 66.12\% (+2.37). However, when all the classification and coordinate prediction tasks are present in one model (A+V+N+G+H), we achieve our best Top1 accuracy at 68.99\% (+5.24) and our best mean class accuracy 61.40\% (+6.05 from STL at 55.35\%). 

\subsection{Comparison to the state-of-the-art}
\label{sec:exp:comp}
In Tables~\ref{tab:3_sota_epic} and \ref{tab:4:egtea_sota_actions} we compare with the state-of-the-art in action recognition for EPIC-Kitchens and EGTEA Gaze+, respectively. For EPIC we demonstrate slightly lower but comparable performance to the top methods for the seen (s1) and unseen (s2) test splits, by requiring only a fraction of the input. For example \cite{furnari_what_2019} requires RGB and flow at test time and \cite{wu_long-term_2019} utilizes knowledge from past video segments, in effect having a larger temporal view of the action. However, we still outperform the attention mechanism of \cite{sudhakaran_lsta_2019}. For EGTEA, we test against several methods, for different metrics. Top1 recognition accuracy for the first split at the clip level is reported in \cite{huang_mutual_2019}  (55.63\%) and in \cite{sudhakaran_attention_2018} (62.17\%) where we improve by 13.36\% and 6.82\% respectively. Li \etal \cite{li_eye_2018} report 47.71\% mean class accuracy on the first split at the clip level (and 53.3\% on the video level). Our method depending on the task combination reaches 58.91\% up to 61.4\% (+11.2 to +13.69 respectively). 

For a more elaborate comparison on EGTEA Gaze+, we train the A+V+N+G+H model for splits 2 and 3 and average the Top1 accuracy over all splits. We achieve 65.7\% Top1 accuracy which is the highest among the reported values by a margin of 3.84\%. For future reference we also report the mean class accuracy averaged over the three splits (57.6\%).

\begin{figure*}
    \centering
    \renewcommand{\arraystretch}{0.3}
    \begin{tabular}{l@{\hspace{2pt}}c@{\hspace{2pt}}c@{\hspace{2pt}}c@{\hspace{2pt}}c@{\hspace{2pt}}c@{\hspace{2pt}}c@{\hspace{2pt}}c@{\hspace{2pt}}c@{\hspace{2pt}}c}
        \raisebox{0.045\linewidth}{(a)} & 
        \includegraphics[width=0.1\linewidth]{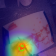} & \includegraphics[width=0.1\linewidth]{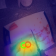} &
        \includegraphics[width=0.1\linewidth]{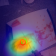} &
        \includegraphics[width=0.1\linewidth]{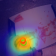} &
        \includegraphics[width=0.1\linewidth]{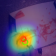} &
        \includegraphics[width=0.1\linewidth]{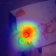} &
        \includegraphics[width=0.1\linewidth]{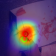} &
        \includegraphics[width=0.1\linewidth]{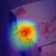} &
        \includegraphics[width=0.1\linewidth]{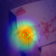}\\
        \raisebox{0.045\linewidth}{(b)} & \includegraphics[width=0.1\linewidth]{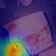} & \includegraphics[width=0.1\linewidth]{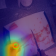} &
        \includegraphics[width=0.1\linewidth]{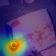} &
        \includegraphics[width=0.1\linewidth]{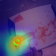} &
        \includegraphics[width=0.1\linewidth]{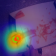} &
        \includegraphics[width=0.1\linewidth]{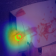} &
        \includegraphics[width=0.1\linewidth]{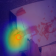} &
        \includegraphics[width=0.1\linewidth]{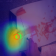} &
        \includegraphics[width=0.1\linewidth]{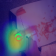}\\
        \raisebox{0.045\linewidth}{(c)} & \includegraphics[width=0.1\linewidth]{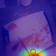} & \includegraphics[width=0.1\linewidth]{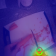} &
        \includegraphics[width=0.1\linewidth]{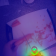} &
        \includegraphics[width=0.1\linewidth]{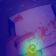} &
        \includegraphics[width=0.1\linewidth]{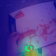} &
        \includegraphics[width=0.1\linewidth]{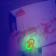} &
        \includegraphics[width=0.1\linewidth]{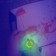} &
        \includegraphics[width=0.1\linewidth]{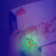} &
        \includegraphics[width=0.1\linewidth]{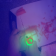}\\
    \end{tabular}
    \caption{(a) Gaze, (b) left hand, (c) right hand coordinate prediction from the A+G+H model. The green circle represents the ground truth coordinate, the red circle the predicted coordinate and the underlying heatmap the 2D probability distribution \(\hat{Z}\).}
    \label{fig3:visuals_g_lh_rh}
\end{figure*}

An additional interesting scope from EGTEA is gaze estimation. Since a number of our models are able to predict gaze on the input frames, we proceed to evaluate it with two standard metrics in the literature: Average Angle Error in degrees (AAE) and Area Under the Curve (AUC) \cite{riche_saliency_2013} following \cite{huang_mutual_2019}. For evaluation we use only the frames from the clips of the first test split for which after resizing and cropping to \(224\times224\) there is a valid ground truth gaze point in this area, regardless of the gaze type. This leads to the evaluation of 177,292/206,649 (85.79\%) frames from 2,022 clips (the remaining frames are not considered). The results are shown in Table~\ref{tab:5:egtea_sota_gaze}. We discover that the gaze estimation techniques which are explicitly designed to model gaze through elaborate attention mechanisms such as \cite{huang_mutual_2019} achieve lower angular error (-3.11\textdegree) although our model (A+G+H) improves over \cite{huang_salicon_2015} and is very close to \cite{li_eye_2018}. Furthermore, considering AUC, our model is second best to \cite{huang_mutual_2019} with a -0.06 margin. The two metrics imply that our method is able to produce gaze predictions that lie in the vicinity of the ground truth (high AUC) but with an angular offset with respect to the exact ground truth gaze. In Figure~\ref{fig3:visuals_g_lh_rh} we show and qualitatively assess gaze and hand predictions. The images show both the predicted saliency in heatmap form as well as its transformation into a single point per frame for gaze and each hand.

\begin{table}[ht]
    \centering
    \begin{tabular}{@{\hspace{0pt}}l@{\hspace{1pt}}|@{\hspace{0pt}}c@{\hspace{3pt}}c@{\hspace{3pt}}c@{\hspace{2pt}}|@{\hspace{0pt}}c@{\hspace{3pt}}c@{\hspace{3pt}}c@{\hspace{0pt}}}
         \multicolumn{1}{l}{} & \multicolumn{3}{c}{Top1 Acc. (\%)} & \multicolumn{3}{c}{Top5 Acc. (\%)}\\
        Method & Actions & Verbs & Nouns & Actions & Verbs & Nouns
        \\
        \hline
        \multicolumn{7}{l}{\textbf{Test s1 (Seen kitchens)}} \\
        TSN \cite{damen_scaling_2018} & 20.54 & 48.23 & 36.71 & 39.79 & 84.09 & 62.32 \\
        LSTA \cite{sudhakaran_lsta_2019} & 30.33 & 59.55 & 38.35 & 49.97 & 85.77 & 61.49 \\ 
        \textit{Ours (all tasks)} & 29.73 & 56.00 & 40.15 & 50.95 & 87.06 & 64.07 \\
        RU \cite{furnari_what_2019} & \textbf{33.06} & 56.93 & 43.05 & \textbf{55.32} & 85.68 & 67.12 \\
        LFB \cite{wu_long-term_2019} & 32.70 & \textbf{60.00} & \textbf{45.00} & 55.30 & \textbf{88.40} & \textbf{71.80} \\
        \hline
        \multicolumn{7}{l}{\textbf{Test s2 (Unseen kitchens)}} \\
        TSN \cite{damen_scaling_2018} & 10.89 & 39.40 & 22.70 & 25.26 & 74.29 & 45.72 \\
        LSTA \cite{sudhakaran_lsta_2019} & 16.63 & 47.32 & 22.16 & 30.39 & 77.02 & 43.15 \\
        \textit{Ours (all tasks)} & 17.86 & 45.99 & 26.25 & 35.68 & \textbf{77.98} & 50.19 \\
        RU \cite{furnari_what_2019} & 19.49 & 43.67 & 26.77 & 37.15 & 73.30 & 48.28 \\
        LFB \cite{wu_long-term_2019} & \textbf{21.20} & \textbf{50.90} & \textbf{31.50} & \textbf{39.40} & 77.60 & \textbf{57.80}
    \end{tabular}
    \caption{Comparison on action recognition against state-of-the-art methods on EPIC-Kitchens. Our method is consistently close to the best performing, while requiring less information at test time.}
    \label{tab:3_sota_epic}
\end{table}

\begin{table}[ht]
    \centering
    \begin{tabular}{l|@{\hspace{3pt}}c@{\hspace{5pt}}c@{\hspace{3pt}}|@{\hspace{3pt}}c@{\hspace{5pt}}c@{\hspace{0pt}}}
        \multicolumn{1}{l}{} & \multicolumn{2}{c}{Split 1} & \multicolumn{2}{c}{Avg. Splits 1-3}\\
        Method & Top1 & Mean Cls & Top1 & Mean Cls \\
        \hline
        Li \etal \cite{li_eye_2018} & - & 47.71  & - & -\\ 
        MCN \cite{huang_mutual_2019} & 55.63 & - & - & - \\ 
        RU \cite{furnari_what_2019} & - & - & 60.20 & -\\ 
        ego-rnn \cite{sudhakaran_attention_2018} & 62.17 & - & 60.76 & - \\
        LSTA \cite{sudhakaran_lsta_2019} & - & - & 61.86 & -\\
        \hline
        \textit{Ours (all tasks)} & \textbf{68.99} & \textbf{61.40} & \textbf{65.70} & \textbf{57.60}
    \end{tabular}
    \caption{Action recognition comparison on EGTEA Gaze+. We compare against the available values from each paper.}
    \label{tab:4:egtea_sota_actions}
\end{table}

\begin{table}
    \centering
    \begin{tabular}{l|c|c}
        Method & AAE & AUC \\
        \hline
        SALICON \cite{huang_salicon_2015} & 11.17 & 0.881 \\
        \textit{Ours (A+G+H)} & \textit{8.90} & \textit{0.926} \\
        Li \etal \cite{li_eye_2018} & 8.58 & 0.87 \\
        DFG \cite{zhang_anticipating_2019}& 6.30 & 0.923 \\
        Huang \etal \cite{huang_predicting_2018} & 6.25 & 0.925 \\
        MCN \cite{huang_mutual_2019} & \textbf{5.79} & \textbf{0.932}
    \end{tabular}
    \caption{Gaze estimation comparison on EGTEA Gaze+ split 1. AAE lower is better, AUC higher is better. SALICON \cite{huang_salicon_2015}, Li \etal \cite{li_eye_2018}, DFG \cite{zhang_anticipating_2019} and Huang \etal \cite{huang_predicting_2018} are reported from \cite{huang_mutual_2019}.}
    \label{tab:5:egtea_sota_gaze}
\end{table}
\vspace{-10pt}

\section{Discussion}
\label{sec:discussion}
Our first aim with MTL is to drive the focus of the activation maps around hand regions and their movements. By training for the hand coordinate task we imply greater importance to them and introduce this requirement to the weights of the shared network block via gradient descent. An example of the expected behavior of the network is in Figure~\ref{fig:1}, where the class activation maps after the last convolution layer cover a larger area of the visible hands. 

The task of gaze prediction is similar to hand detection in that it expects the network to focus on specific regions of the input frames. The difference is that these regions do not necessarily contain the well-structured form of hands, but the salient areas of a scene, which are not predetermined. This limits the ability of region-specific features to become significant making it a dataset- and class-specific quality. 

In both datasets, we observe almost consistent improvements over STL with the introduction of hands and other tasks. However, the choice of tasks involves a significant amount of intuition as well as the weighing of their importance in the loss function. In this work, we use a naive weighing mechanism and consider all tasks equal regardless of the loss they incur. When training multitask models for EPIC-Kitchens we notice high values of loss in the classification tasks, which stem from the class imbalance and the large number of action, verb and noun classes. These losses initially affect their individual layers, but the backpropagated gradients to the shared weights are also higher, affecting the representation in an unbalanced way. In the EPIC-Kitchens results we see that by adding classification tasks with more classes (such as N at V+N+H, or A at A+V+N+H), we get a worse verb classifier. This is caused by the high losses incurred from the added tasks. In certain cases, they act as regularization but when they are too high they can increase training times and even prevent convergence. We believe further research is needed in MTL for video recognition to establish weighing mechanisms such as \cite{sener_multi-task_2018} for a more optimized shared parameter space.

On the other hand, on EGTEA Gaze+, MTL consistently outperforms STL for every task combination. This shows that carefully designing the classification tasks (e.g. fewer classes, balanced dataset) can be mutually beneficial to all, but more importantly to the action task we are most interested in. Incorporating hands in training confirms our initial intuition that motions create a fitting side-task to actions, increasing performance. A possible reason for the higher improvement due to hands on EGTEA compared to EPIC is the presence of hand annotations from the former in the training set of the hand detector of \cite{kapidis_egocentric_2019}. This, could result in a more accurate synthetic hand dataset for EGTEA. Finally, improvements due to gaze validate the connection between actions and gaze \cite{huang_mutual_2019,li_eye_2018} also from the perspective of MTL. 

A possible pitfall of MTL is competition among tasks with negative effects on performance. This is possible if tasks are incompatible or if the network is not large enough to create a representation that engulfs the different aspects of information required for each one. The former regards a (lack of) conceptual relevance, for example verbs and nouns on EPIC, or structural, for example classification layers operating differently from coordinate regression ones and possibly requiring a distinct representation in earlier layers. The concept of task compatibility has been studied for other domains in \cite{caruana_multitask_1997} concluding that the degree of assistance from an extra task in learning another cannot be clear a priori without experimentation. This can be viewed as treating tasks as an additional hyperparameter. 
An example of task incompatibility with respect to actions is when both gaze and hands are used for action recognition (A+G+H) but lead to worse performance than training individually (A+G, A+H). Adding a task may not improve as much as another combination but can also reduce the expected baseline performance. Here, however, the trade-off is actions and hands contributing towards an optimal gaze detector.

\section{Conclusions}
\label{sec:conclusions}

In this work, we have developed a Multitask Learning scheme for egocentric action recognition that supports a variable number of tasks. We train for actions together with related classification tasks, such as verbs and nouns, and show that performance on one or all of them will improve over the single-task baseline. We further combine classification with coordinate regression tasks to learn the egocentric left and right hand and gaze locations, by predicting coordinate sequences for video segments exploiting the temporal dimension of 3D CNNs. We highlight that having a network estimate coordinates allows it to focus more on certain areas of the activation maps with higher correspondence to the hands or other salient objects in the original image. Our tests on the EPIC-Kitchens show improvements on action recognition performance over single-task. On EGTEA Gaze+ we achieve state-of-the-art performance in action recognition reaching 65.70\% surpassing the previous best by more than 3.8\%. In addition, with multitask learning we can produce accurate hand detectors, as well as gaze predictors with performance close to state-of-the-art.

\section*{Acknowledgment}
This project has received funding from the EU H2020 research and innovation program under the Marie Sk\l{}odowska Curie grant agreement No 676157 (ACROSSING).

{\small
\bibliographystyle{ieee_fullname}
\bibliography{ICCVW2019.bib}
}

\end{document}